\documentclass[]{fairmeta}

\usepackage{wrapfig}
\usepackage{tabularx}
\usepackage{textcomp}
\usepackage{stfloats}
\usepackage{url}
\usepackage{verbatim}
\usepackage{titlesec}
\usepackage{tocloft}
\usepackage{adjustbox}
\usepackage{tikz}
\usepackage{comment}
\usepackage{amsmath,amsfonts,amssymb}
\usepackage{colortbl}
\usepackage{color}
\usepackage{hyperref}
\usepackage{graphicx}
\usepackage{subcaption} 
\usepackage{multirow}
\usepackage{booktabs}
\usepackage{xcolor}
\usepackage{pifont}
\usepackage{bbding}
\usepackage{fontawesome}
\usepackage{xspace}
\usepackage{float}
\usepackage{algorithm}
\usepackage{listings}
\usepackage{algcompatible}
\usepackage{algpseudocode}
\usepackage{soul}
\usepackage{array}
\usepackage{tcolorbox}
\usepackage{diagbox}
\usepackage{makecell}
\usepackage{rotating}
\usepackage{enumitem}

\usepackage{amsmath,amsfonts,bm}

\def\eqref#1{equation~\ref{#1}}

\def\1{\bm{1}}

\DeclareMathAlphabet{\mathsfit}{\encodingdefault}{\sfdefault}{m}{sl}
\SetMathAlphabet{\mathsfit}{bold}{\encodingdefault}{\sfdefault}{bx}{n}

\RequirePackage{xspace}
\makeatletter
\DeclareRobustCommand\onedot{\futurelet\@let@token\@onedot}
\def\@onedot{\ifx\@let@token.\else.\null\fi\xspace}

\def\eg{\emph{e.g}\onedot} 
\def\ie{\emph{i.e}\onedot}

\makeatother

\definecolor{adptorange}{RGB}{248, 205, 172}
\definecolor{cmpblue}{RGB}{189, 215, 238}
\definecolor{cmpblue}{RGB}{189, 215, 238}

\definecolor{our_red}{RGB}{232,157,160}
\definecolor{our_blue}{RGB}{136,206,230}
\definecolor{our_orange}{RGB}{246,200,168}
\definecolor{our_green}{RGB}{178,211,164}

\definecolor{attn_code0}{RGB}{247,215,200}
\definecolor{attn_code1}{RGB}{238,169,139}
\definecolor{mlp_code0}{RGB}{204,201,221}
\definecolor{mlp_code1}{RGB}{102,95,153}

\definecolor{token_blue}{RGB}{84, 120, 140}

\newlength\savewidth

\newcolumntype{x}[1]{>{\centering\arraybackslash}p{#1pt}}
\newcolumntype{y}[1]{>{\raggedright\arraybackslash}p{#1pt}}
\newcolumntype{z}[1]{>{\raggedleft\arraybackslash}p{#1pt}}

\renewcommand{\paragraph}[1]{\vspace{1mm}\noindent\textbf{#1}}

\renewcommand{\paragraph}[1]{\vspace{1.25mm}\noindent\textbf{#1}}

\definecolor{codeblue}{rgb}{0.25, 0.5, 0.5}
\definecolor{codekw}{rgb}{0.35, 0.35, 0.75}
\lstdefinestyle{Pytorch}{
    language = Python,
    backgroundcolor = \color{white},
    basicstyle = \fontsize{9pt}{8pt}\selectfont\ttfamily\bfseries,
    columns = fullflexible,
    aboveskip=1pt,
    belowskip=1pt,
    breaklines = true,
    captionpos = b,
    commentstyle = \color{codeblue},
    keywordstyle = \color{codekw},
}

\definecolor{green}{HTML}{009000}
\definecolor{red}{HTML}{ea4335}

\newcommand{\eqcontrib}{\clubsuit}
\newcommand{\correspond}{\spadesuit}

\newcommand{\ours}{\textsc{VL-Cogito}\xspace}

\newtcolorbox{promptblock}{
    colback=gray!5,
    colframe=gray!15,
    boxrule=0.5pt,
    arc=3pt,
    left=12pt,
    right=12pt,
    top=8pt,
    bottom=8pt,
    boxsep=8pt,
    breakable
}

\title{\ours: Progressive Curriculum Reinforcement Learning for Advanced Multimodal Reasoning}

\author[1,2,3,\eqcontrib]{Ruifeng Yuan}
\author[1,\eqcontrib]{Chenghao Xiao}
\author[1]{Sicong Leng}
\author[1]{Jianyu Wang}
\author[1]{Long Li}
\author[1]{Weiwen Xu}
\author[1]{Hou Pong Chan}
\author[1,2]{Deli Zhao}
\author[1,2]{Tingyang Xu}
\author[3]{Zhongyu Wei}
\author[1,\correspond]{Hao Zhang}
\author[1,2]{Yu Rong}

\affiliation{$^1$DAMO Academy, Alibaba Group}
\affiliation{$^2$Hupan Lab}
\affiliation{$^3$Fudan University}

\contribution[\eqcontrib]{Equal Contribution}
\contribution[\correspond]{Corresponding Author}

\abstract{
Reinforcement learning has proven its effectiveness in enhancing the reasoning capabilities of large language models. Recent research efforts have progressively extended this paradigm to multimodal reasoning tasks. Due to the inherent complexity and diversity of multimodal tasks, especially in semantic content and problem formulations, existing models often exhibit unstable performance across various domains and difficulty levels. To address these limitations, we propose \ours, an advanced multimodal reasoning model trained via a novel multi-stage Progressive Curriculum Reinforcement Learning (PCuRL) framework. PCuRL systematically guides the model through tasks of gradually increasing difficulty, substantially improving its reasoning abilities across diverse multimodal contexts. The framework introduces two key innovations: (1) an online difficulty soft weighting mechanism, dynamically adjusting training difficulty across successive RL training stages; and (2) a dynamic length reward mechanism, which encourages the model to adaptively regulate its reasoning path length according to task complexity, thus balancing reasoning efficiency with correctness. Experimental evaluations demonstrate that \ours consistently matches or surpasses existing reasoning-oriented models across mainstream multimodal benchmarks spanning mathematics, science, logic, and general understanding, validating the effectiveness of our approach.
}

\date{\today}
\correspondence{Hao Zhang, \email{hz.hhea2e@alibaba-inc.com}}
\metadata[Homepage]{\url{https://github.com/alibaba-damo-academy/VL-Cogito}}

\begin{document}
\thispagestyle{firstheader}
\maketitle
\pagestyle{fancy}
\fancyhf{}
\fancyfoot[C]{\thepage}

\section{Introduction}
\label{sec:intro}
Recent advancements of Reinforcement Learning (RL) in Large Language Models (LLMs), such as OpenAI o-series~\citep{openai2024openaio1}, Kimi-K1.5~\citep{kimiteam2025kimik15}, and DeepSeek-R1~\citep{deepseekai2025deepseekr1}, highlight its promise to incentivize the long-chain reasoning abilities of LLMs, enabling them to effectively tackle complex tasks involving code generation, mathematical problems, and scientific reasoning. In particular, RL with verifiable rewards, such as GRPO~\citep{shao2024deepseekmath}, has emerged as a pivotal paradigm for training a ``slow-thinking'' system, which directly employs rule-based rewards and promotes the generation and iterative refinement of multiple reasoning paths, using group-based relative advantage estimations. This approach has notably improved the performance of LLMs to solve challenging reasoning tasks, pushing the boundaries of what LLMs can achieve in various domains~\citep{yeo2025demystifying,chen2025reasoning}.

Following the success of this training paradigm in LLMs, researchers have increasingly explored its application to Multimodal Large Language Models (MLLMs). Initial efforts primarily focus on adapting these techniques to specific multimodal domains such as mathematics and logic~\citep{meng2025mm,huang2025vision,chen2025mathflow}. As MLLMs are not constrained to textual modalities, it has the opportunity to enable reasoning across diverse domains. The spectrum ranges from straightforward chart interpretation~\citep{huang-etal-2024-chart}, complex geometry problems~\citep{lu2024mathvista}, to intricate scientific analysis~\citep{lu2022scienceqa}. With this expanded scope of domains, the heterogeneity among different problem types becomes increasingly apparent, presenting a challenge in effectively learning reasoning skills for tasks of varying complexities and types~\citep{wang2025multimodal,bi2025reasoning,li2025perception}.

In this work, we introduce \ours, a reasoning-oriented MLLM trained on an extensive dataset comprising diverse multimodal task domains.
To address the varying levels of data difficulty, we propose a \textbf{P}rogressive \textbf{Cu}rriculum \textbf{R}einforcement \textbf{L}earning framework (\textbf{PCuRL}). Central to our approach is a novel curriculum learning strategy, which systematically guides the model through progressively complex tasks to build robust reasoning capabilities. Specifically, we introduce an \textit{online difficulty soft weighting mechanism} to dynamically adjust data selection according to predefined difficulty distributions across successive training stages, allowing the model to incrementally transition from mastering simpler questions to effectively handling intricate problems. Moreover, another notable innovation in PCuRL is the \textit{dynamic length reward mechanism}. Conventional reasoning models often extend reasoning length indiscriminately, potentially compromising efficiency. In contrast, our dynamic reward explicitly incentivizes the model to adapt its reasoning length based on the demands of individual problems. This adaptive strategy ensures the model not only engages in thorough reasoning for complex tasks but also maintains efficiency and succinctness when simpler problems arise, thus optimizing performance across diverse scenarios.

To thoroughly assess the efficacy of PCuRL and the performance of \ours, we perform extensive experiments on various multimodal reasoning benchmarks spanning mathematical, scientific, and general domains. It is worth noting that \ours bypasses a cold-start SFT phase and is instead trained directly from the backbone model via PCuRL using the GRPO approach. The results demonstrate that \ours achieves state-of-the-art or highly competitive performance across all evaluation sets, underscoring the superiority and effectiveness of our proposed method. Furthermore, comprehensive ablation studies are conducted to analyze the contribution of each module, confirming their respective importance. Visualizations of the training process and detailed case studies validate that the progressive curriculum strategy ensures training stability while enhancing both effectiveness and efficiency.

Our contributions can be summarized as follows:
\begin{itemize}
    \item We propose PCuRL, a novel multi-stage progressive curriculum RL framework, incorporating an online difficulty soft weighting mechanism that progressively exposes models to increasingly challenging tasks, enhancing multimodal reasoning capabilities. In addition, we introduce a dynamic length reward mechanism designed to encourage the model to modulate reasoning length based on question-specific complexity, effectively balancing depth and efficiency.

    \item \ours achieves the state-of-the-art or highly competitive performance in multimodal benchmarks across various domains, underscoring the effectiveness and versatility of our proposed framework and training pipeline.

    \item Extensive ablation studies confirm that our progressive curriculum learning mechanism effectively and consistently increases reasoning depth, leading to improved performance on complex tasks. The dynamic length reward strategy allows the model to produce concise answers for simple problems while promoting longer, more in-depth reasoning for more difficult ones. Overall, PCuRL delivers substantial and balanced gains in accuracy, training stability, and efficiency across a range of task difficulties.
\end{itemize}

\section{Related Work}
\label{sec:related_work}
As RL-based reasoning models have demonstrated success in LLMs, significant research attention has shifted toward adapting these advances to multimodal large language models (MLLMs; \citet{yao2024mulberry,xu2025llavacot,wang2025visualprm,peng2025skyworkr1v,xia2025visionaryr1}). Some work aims to improve MLLMs' reasoning abilities directly via RLVR-style optimization~\citep{kimi2025kimivl,wang2025skyworkr1v2,guo2025seed15vl,kwaikeye2025kwaikeyevl}. For instance, Vision-R1~\citep{huang2025vision} distills long-chain reasoning data from existing models using modality bridging and introduces progressive thinking suppression to address overthinking. R1-OneVision~\citep{yang2025r1} presents a model that converts visual information into formal textual representations for accurate cross-modal reasoning, while OpenVLThinker~\citep{deng2025openvlthinker} combines supervised fine-tuning with reinforcement learning in an iterative training framework.

To enhance the stability and effectiveness of RLVR training, some research investigates advanced strategies for multimodal reasoning models, such as data selection~\citep{wang2025vl,meng2025mm}, reward design~\citep{shen2025vlmr1,tan2025reasonrft}, and advanced RL recipes~\citep{peng2025lmmr1,zhang2025r1,liu2025othinkmr1,yao2025r1sharevl,chen2025grpocare}. MM-Eureka~\citep{meng2025mm} demonstrates that difficulty-based data selection is critical for successful RL training. ThinkLite-VL~\citep{wang2025sota} proposes a data-efficient visual reasoning framework employing Monte Carlo Tree Search for sample difficulty estimation, enabling RL with significantly reduced training data. MMR1~\citep{leng2025mmr1} highlights the gradient vanishing issue in GRPO~\citep{razin2025what} and introduces a variance-aware sampling strategy to mitigate it. R1-VL~\citep{zhang2025r1} presents StepGRPO, a multi-stage online RL framework that incorporates step-wise reasoning rewards to improve the reasoning capabilities of MLLMs. VL-Rethinker~\citep{wang2025vl} introduces selective sample replay to address vanishing advantages and a forced rethinking mechanism to promote slow and deliberate reasoning. Praxis-VLM~\citep{hu2025praxisvlm} proposes an adaptive R1 reward that targets different skills (formatting and reasoning) of MLLMs at different RL training stages. 
In contrast, our work proposes a novel RL recipe inspired by curriculum learning. We integrate online difficulty soft weighting to train the model on tasks of increasing difficulty, and introduce a novel dynamic length reward design that adaptively determines the ideal reasoning length for each prompt.

\section{Preliminaries}
\label{sec:preliminaries}

\begin{figure}[t]
    \centering
    \includegraphics[width=\textwidth]{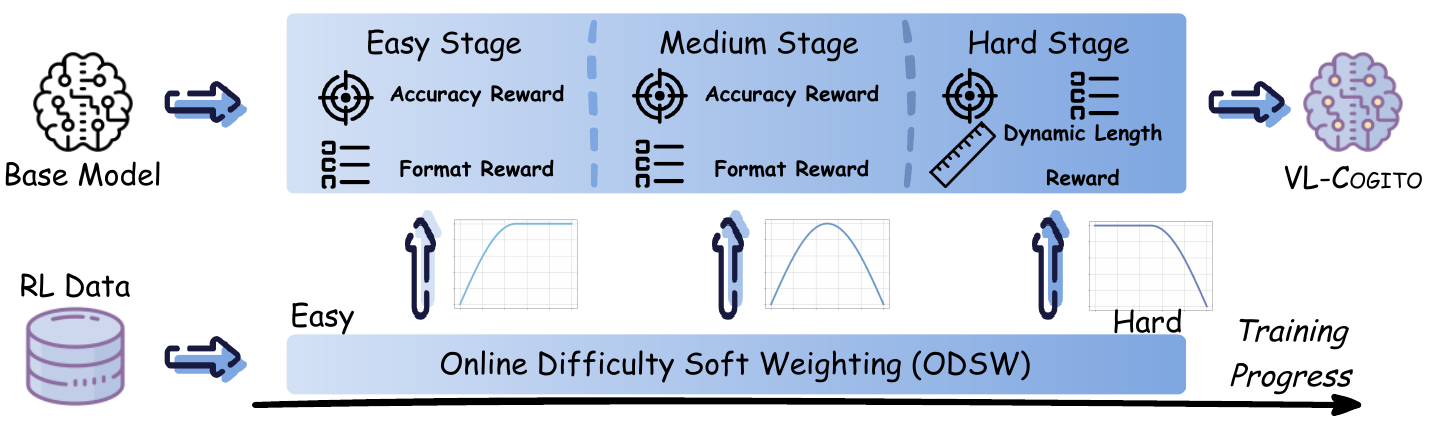}
    \caption{An overview of the proposed Progressive Curriculum Reinforcement Learning (PCuRL) framework. It consists of two key components: (1) a multi-stage curriculum RL structure that utilizes online difficulty soft weighting, which partitions the training progress into different stages based on task difficulty; (2) a dynamic length reward mechanism that encourages the model to adapt its reasoning chain length according to task complexity, rather than indiscriminately increasing it. In the Easy stage, the model tends to assign higher weights to relatively easier questions for policy optimization, a pattern that similarly applies to the Medium and Hard stages.}
    \label{fig:pipeline}
\end{figure}

Group Relative Policy Optimization (GRPO; \citet{shao2024deepseekmath}) is a reinforcement learning algorithm designed to improve the efficiency and effectiveness of training large language models. It estimates the advantages of language model generations by comparing responses within a group specific to the same input. Given an input \( x \), the behavior policy \( \pi_{\theta_{\text{old}}} \) first samples a group of \( G \) candidate responses \(\{y_i\}_{i=1}^{G}\). At time step \( t \), the advantage for the \( i \)-th response is calculated based on the following equation:
\begin{equation}
{A}_{i,t} = \frac{r(x, y_i) - \text{mean}(\{r(x, y_1), \ldots, r(x, y_G)\})}{\text{std}(\{r(x, y_1), \ldots, r(x, y_G)\})}.
\end{equation}

The GRPO also adopts a clipped surrogate objective like PPO~\citep{schulman2017ppo}:
\begin{equation}
\frac{1}{G} \sum_{i=1}^{G} \frac{1}{|y_i|} \sum_{t=1}^{|y_i|} \min \left[ \frac{\pi_\theta(y_{i,t} \mid x, y_{i,<t})}{\pi_{\theta_{\text{old}}}(y_{i,t} \mid x, y_{i,<t})} {A}_{i,t}, \text{clip} \left( \frac{\pi_\theta(y_{i,t} \mid x, y_{i,<t})}{\pi_{\theta_{\text{old}}}(y_{i,t} \mid x, y_{i,<t})}, 1 - \epsilon, 1 + \epsilon \right) {A}_{i,t} \right],
\end{equation}
where \( \epsilon \) serves as a hyperparameter that controls the tolerance for policy deviation. The \text{clip} function is used to prevent excessively large updates by maintaining the ratio between the current policy and the reference policy within a specified range.

Reinforcement Learning with Verifiable Rewards (RLVR) has become a major paradigm for training reasoning models in domains like mathematical reasoning, exemplified by OpenAI's o-series~\citep{openai2024openaio1,openai2025o3} and DeepSeek-R1~\citep{deepseekai2025deepseekr1}. The key idea is to provide the model with clear and objective feedback based on predefined correctness criteria. RLVR utilizes rule-based functions to evaluate the accuracy of a model's output. A format reward is often combined with the accuracy reward to distinguish between the reasoning process and the final answer:
\begin{equation}
r(x, y_i) = r_{\text{acc}}(x, y_i) + r_{\text{format}}(y_i),
\end{equation}
where the $r_{\text{acc}}$ refers to a binary accuracy reward of 1 for correct and 0 for incorrect outputs, and $r_{\text{format}}$ is a strict binary format reward used to constrain the structure of the model's output.

\section{Methodology}
\label{sec:methodology}

\subsection{Data Curation}
\label{ssec:data_curation}
To promote robust performance and strong generalization across a variety of domains, we curated an extensive collection of open-source multimodal datasets, covering $23$ datasets distributed across six distinct task categories. 
(1) \textbf{Mathematical Reasoning}: Geometry3K training set~\citep{geometry3k}, GeoQA+~\citep{cao2022gpqaplus}, Geos~\citep{seo2015solving}, GeomVerse~\citep{kazemi2024geomverse}, Inter-GPS~\citep{lu2021inter}, MultiMath~\citep{peng2024multimath}; 
(2) \textbf{Logical Reasoning}: Raven~\citep{mm_raven}, MM-IQ~\citep{cai2025mmiq}, EasyArc~\citep{unsal2025easyarc}; 
(3)\textbf{Counting}: CLEVR-Math~\citep{lindström2022clevrmath}, Super-CLEVR~\citep{li2023super};
(4) \textbf{Science Reasoning}: AI2D training set~\citep{kembhavi2016ai2d}, ScienceQA training set~\citep{lu2022scienceqa}, TQA~\citep{kembhavi2017tqa});
(5) \textbf{Chart Understanding}: ChartQA training set~\citep{masry2022chartqa}, TabMWP~\citep{lu2023tabmwp}, DVQA~\citep{kafle2018dvqa}, FigureQA~\citep{kahou2018figureqa}, ArXivQA~\citep{li2024arxivqa}, InfographicVQA~\citep{mathew2022infographicvqa};
and (6) \textbf{General Image Understanding}: OKVQA~\citep{marino2019okvqa}, VQA2.0~\citep{antol2015vqa}, LLaVA-CoT~\citep{xu2025llavacot}.

For the data used in reinforcement learning, we selectively sample a subset of the data based on our designed quality and category criteria to construct the training set. During data curation, our primary objective is to enhance the overall difficulty and coverage of the training samples, encouraging the model to perform more in-depth reasoning. To this end, two additional measures are implemented: (1) \textbf{Open-ended Format}: To prevent the reasoning model from relying on superficial cues present in specific answer formats, such as multiple-choice, we reformulate most samples into an open-ended QA format; (2) \textbf{Difficulty Sampling}: To exclude questions that do not necessitate genuine reasoning, we employ difficulty-based sampling approach by removing samples that achieve above $50\%$ accuracy over $8$ trails using Qwen2.5-VL-7B-Instruct~\citep{bai2025qwen25vl}. The resultant distribution of RL training data following these filtering procedures is presented in Table~\ref{tab:datasets} of Appendix~\ref{apdx:sec:data}.

\subsection{Progressive Curriculum Reinforcement Learning Framework}
Consistent with prior work~\citep{meng2025mm,wang2025vl,wang2025sota}, we leverage the Group Relative Policy Optimization (GRPO)-based RLVR scheme. On this foundation, we propose a \underline{P}rogressive \underline{Cu}rriculum \underline{R}einforcement \underline{L}earning (\textbf{PCuRL}) framework, which incrementally trains the model on tasks of increasing difficulty. The goal is to enable the MLLM to gradually acquire more sophisticated reasoning skills and extend its reasoning chain, while still maintaining concise and effective responses on simpler tasks. The PCuRL framework comprises two key components: (1) a multi-stage curriculum RL structure that utilizes \textit{online difficulty soft weighting}, and (2) a \textit{dynamic length reward mechanism} that encourages the model to adapt its reasoning chain length according to task complexity, rather than indiscriminately increasing it. The overall architecture of our framework is depicted in Figure~\ref{fig:pipeline}.

\subsubsection{Online Difficulty Soft weighting (ODSW)}
\label{sssec:soft_weight_difficulty}
The difficulty soft weighting mechanism is designed to prioritize target prompts with specific difficulty levels during the RL training process. 
Prior work~\citep{bae2025online,cui2025process} proposes an online hard-weighted difficulty filtering method that discards prompts outside the appropriate difficulty range from each training batch. In contrast, our mechanism allows for more prompts to be considered during training, assigning each a weight that represents its relative importance. Drawing inspiration from ADORA~\citep{gui2025adora}, we realize \textbf{O}nline \textbf{D}ifficulty \textbf{S}oft \textbf{W}eighting (\textbf{ODSW}) by adjusting the advantage values at the prompt level according to the weights that reflect the difficulty of each question prompt. Consequently, prompts with higher weights contribute more substantially to the gradient updates during training. Specifically, these weights are calculated based on the rollout accuracy of each prompt using a predefined function $F$ as:

\begin{figure}[t]
    \centering
    \begin{subfigure}[t]{0.26\textwidth}
        \centering
        \includegraphics[width=\textwidth]{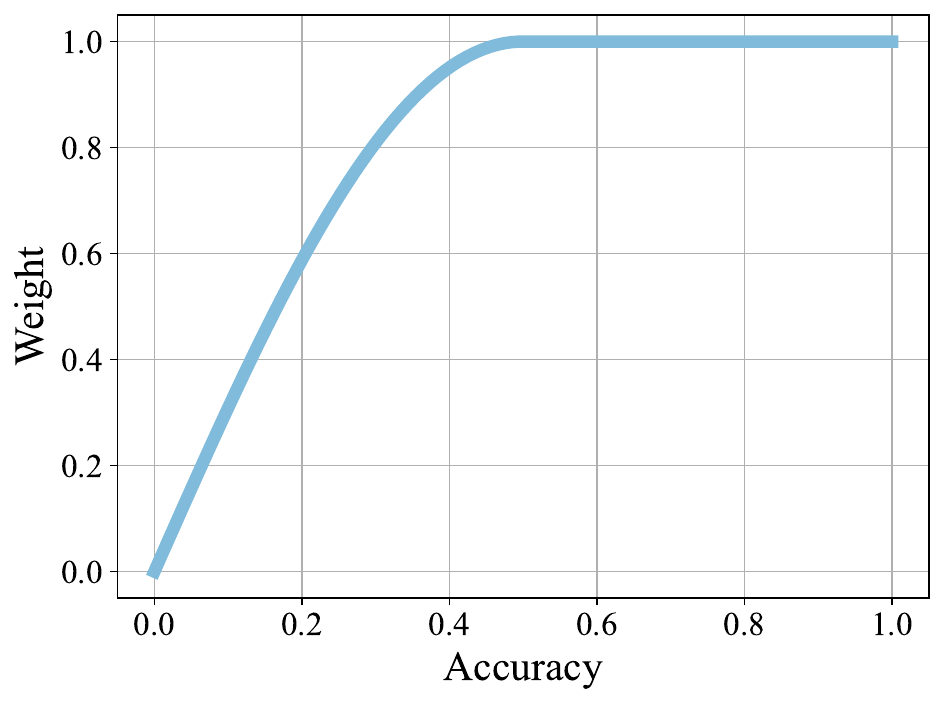}
        \caption{ODSW Easy}
        \label{fig:ODSW_easy}
    \end{subfigure}
    \hfill
    \begin{subfigure}[t]{0.26\textwidth}
        \centering
        \includegraphics[width=\textwidth]{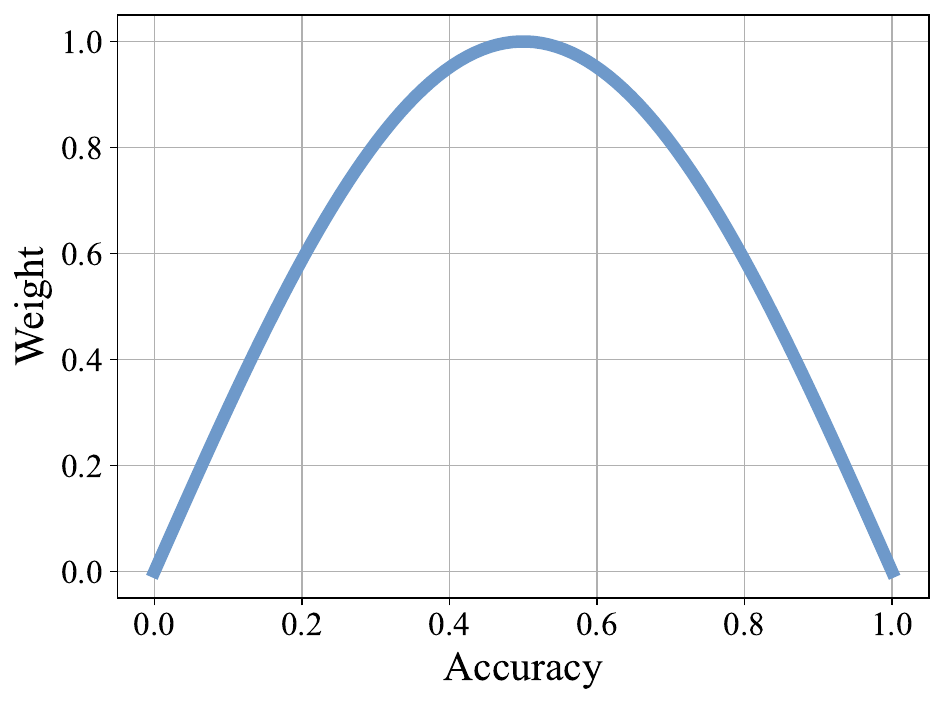}
        \caption{ODSW Medium}
        \label{fig:ODSW_medium}
    \end{subfigure}
    \hfill
    \begin{subfigure}[t]{0.26\textwidth}
        \centering
        \includegraphics[width=\textwidth]{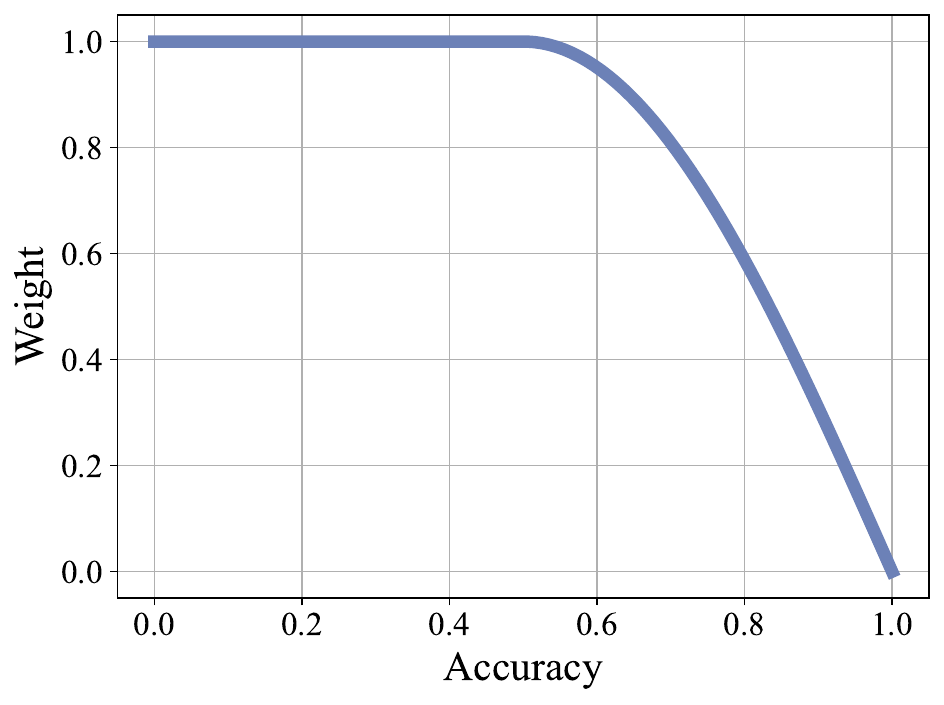}
        \caption{ODSW Hard}
        \label{fig:ODSW_hard}
    \end{subfigure}
    \caption{Three difficulty distributions, \ie, easy, medium, and hard, for the Online Difficulty Soft weighting (ODSW).}
    \label{fig:difficulty}
\end{figure}

\begin{equation}
\hat{A}_{i,t} = F\Big(\frac{1}{G} \sum_{i=1}^{G} \text{acc}(x, y_i)\Big)\cdot{A}_{i,t},
\label{eq:new_advantage}
\end{equation}
where $G$ denotes the number of rollout responses per prompt, $\text{acc}$ indicates whether a rollout is correct or not (\ie, $1$ for correct and $0$ for incorrect), and the definition of $F$ is based on the theory of learnability~\citep{foster2025learning,rutherford2024no,tzannetos2023proximal}.

According to the theory of learnability, prompts that achieve a rollout accuracy close to $0.5$ are considered optimal for RL training, as they present an appropriate level of challenge for effective model learning. Motivated by this theoretical insight, we construct $F$ as a \textit{continuous piecewise function}, which integrates both sine and constant components with $0.5$ as the threshold. The constant components allow the model to focus on the target difficulty, while the sine component ensures a smooth transition between different difficulty levels. This design emphasizes the prompts with the highest learnability, and guides the gradient update direction to explore a specific difficult level. Additionally, it prevents the model from entirely disregarding prompts of other difficulties. $F$ is allowed to flexibly accommodate varying difficulty distributions, facilitating the targeted weighting of prompts that maximize training efficacy. In particular, we design three distinct versions of the function $F$ (as illustrated in Figure~\ref{fig:difficulty}), each corresponding to prompts with predominantly easy, medium, or hard difficulty levels. To provide a baseline for comparison, we also implement a binary weighting strategy alongside our proposed difficulty soft weighting approach. The definitions of different difficulty soft weighting variants and binary weighting strategies are computed as follows:

\begin{itemize}
    \item ODSW Easy (see Figure~\ref{fig:ODSW_easy}): 
    \[
    F(\mathrm{Acc}) = 
    \begin{cases}
    \sin(\pi\cdot\mathrm{Acc}), & \text{if } \mathrm{Acc} \in [0, 0.5) \\
    1, & \text{if } \mathrm{Acc} \in [0.5, 1]
    \end{cases};
    \]
    \item ODSW Medium (see Figure~\ref{fig:ODSW_medium}): 
    \[
    F(\mathrm{Acc})=Sin(\pi\cdot \mathrm{Acc});
    \]
    
    \item ODSW Hard (see Figure~\ref{fig:ODSW_hard}):
    \[
    F(\mathrm{Acc}) = 
    \begin{cases}
    1, & \text{if } \mathrm{Acc} \in [0, 0.5] \\
    \sin(\pi\cdot\mathrm{Acc}), & \text{if } \mathrm{Acc} \in (0.5, 1]
    \end{cases};
    \]
    
    \item Binary Weighting:
    \[
    F(\mathrm{Acc}) =
    \begin{cases}
    1, & \text{if } \mathrm{Acc} \in [T_{\min}, T_{\max}] \\
    0, & \text{otherwise}
    \end{cases}.
    \]
\end{itemize}

For simplicity, we denote the average rollout accuracy as $\mathrm{Acc} = \frac{1}{G} \sum_{i=1}^{G} \text{acc}(x, y_i)$. The $T_{\text{min}}$ and $T_{\text{max}}$ in the binary weighting strategy are predefined ranges of average accuracy.

By integrating explicit difficulty control, \ie, ODSW, with curriculum learning, we facilitate progressive model optimization. Initially, focusing on simpler problems enables the model to efficiently acquire correct reasoning patterns and receive clearer optimization signals, thereby stabilizing the training process. Subsequently, focusing on more challenging problems promotes deeper reasoning and exploration of diverse reasoning paths, further improving performance.

\subsubsection{Dynamic Length Reward (DyLR)} 
Extending the reasoning process typically provides reasoning models with greater space for thought and allows them to explore more diverse reasoning patterns to tackle challenging problems. A common practice is to implement a length reward mechanism, \eg, Cosine Reward~\citep{yeo2025demystifying}, which incentivizes the model to generate longer reasoning paths. However, such a length reward has notable limitations, as it sets a uniform target length and reward strategy across all prompts, resulting in homogeneous reasoning-length adjustments despite variations among tasks. In multimodal reasoning contexts, the preferred reasoning length often differs significantly across tasks. For instance, chart understanding, which primarily relies on perceptual skills, generally necessitates shorter reasoning lengths compared to more complex mathematical tasks, such as geometry problems. Applying a uniform length reward across all tasks without differentiation can thus lead to excessively verbose reasoning in certain cases, while inadequately addressing length requirements in others. Consequently, this indiscriminate approach may impair both reasoning efficiency and model performance.

To address this issue, we propose a \textbf{Dy}namic \textbf{L}ength \textbf{R}eward (\textbf{DyLR}) mechanism to determine the ideal reasoning length for each prompt. The core idea is to estimate an appropriate reasoning length for each prompt by utilizing its associated rollout samples at the prompt level. To be specific, the target reasoning budget for each prompt is defined as the average length of all correct responses within its rollout samples. This strategy encourages the model to generate reasoning paths that closely align with the prompt-specific target length. As training progresses and model accuracy improves, the target lengths adapt dynamically and eventually stabilize. For prompts without any correct responses, the dynamic length reward incentivizes the model to extend reasoning up to a predefined maximum length. The formal definition of the dynamic length reward, $r_{\text{len}}$, is written as:
\begin{equation}
r_{\text{len}}(y_i, y) = 
\begin{cases} 
\text{CosFn}(L_{i}, L_{\text{\color{green}{avg}}}, r_{\text{len}}^{\text{min}}, r_{\text{len}}^{\text{max}}) & \text{if } \text{Acc} > 0 \\ 
\text{CosFn}(L_{i}, L_{\color{red}{\text{max}}}, r_{\text{len}}^{\text{min}}, r_{\text{len}}^{\text{max}}) & \text{if } \text{Acc} = 0
\end{cases},
\end{equation}
where $L_{i}$ refers the reasoning length of $i$-th response $y_i$ of a prompt $x$, $L_{\text{\color{green}{avg}}}$ denotes the average length of all correct responses of the prompt $x$, $L_{\color{red}{\text{max}}}$ represents the preset maximum target length, $r_{\text{len}}^{\text{min}}$ and $r_{\text{len}}^{\text{max}}$ are the preset minimum and maximum length rewards, respectively. The function \text{CosFn} is defined as:
\begin{equation}
\text{CosFn}(L_{i}, L_{\text{tgt}}, r_{\text{len}}^{\text{min}}, r_{\text{len}}^{\text{max}}) = r_{\text{len}}^{\text{min}} + \frac{1}{2}(r_{\text{len}}^{\text{max}} - r_{\text{len}}^{\text{min}})\left(1 - \cos\left(\frac{L_{i}\cdot\pi}{L_{\text{tgt}}}\right)\right),
\end{equation}
where $L_{\text{tgt}}$ is the target reasoning length, \ie, $L_{\text{\color{green}{avg}}}$ or $L_{\color{red}{\text{max}}}$ for $ \text{Acc} > 0$ and $ \text{Acc} = 0$ respectively. Through combining the correctness, format, and dynamic length rewards, the overall reward of PCuRL is defined as:

\begin{equation}
r(x, y_i) = \alpha \cdot r_{\text{acc}}(x, y_i) + \beta \cdot r_{\text{format}}(y_i) + \gamma \cdot r_{\text{len}}(y_i, y),
\end{equation}
where $\alpha$, $\beta$, and $\gamma$ are hyperparameters to control the contributions of each specific reward term.

To mitigate excessive growth in reasoning length that may degrade reasoning quality, we introduce an additional hyperparameter $w$ to the advantages of prompts with zero accuracy, \ie, $\text{Acc} = 0$. Accordingly, when integrating dynamic length reward with difficulty soft weighting, Equation~\ref{eq:new_advantage} is reformulated as:
\begin{equation}
\hat{A}_{i,t} = 
\begin{cases} 
F\Big(\frac{1}{G} \sum_{i=1}^{G} \text{acc}(x, y_i)\Big)\cdot{A}_{i,t} & \text{if } \text{Acc} > 0\\ 
w\cdot F\Big(\frac{1}{G} \sum_{i=1}^{G} \text{acc}(x, y_i)\Big)\cdot{A}_{i,t} & \text{if } \text{Acc} = 0
\end{cases}.
\end{equation}
In general, the incorporation of DyLR enables the model to dynamically adjust its output length rather than conforming to a fixed target, thereby mitigating potential over- or under-thinking issues. For simple tasks, the model generates concise responses through rapid reasoning, while for more complex tasks, it automatically adopts longer reasoning trajectories to derive accurate answers.

\subsubsection{Progressive Curriculum RL}
Based on carefully designed online difficulty soft weighting and dynamic length reward mechanisms, we employ a progressive curriculum learning strategy to systematically guide the model's learning trajectory during RL training, which facilitates a gradual transition from simple tasks to increasingly complex and challenging ones. Similar approaches are widely applied in LLM training to enhance the stability of learning and enable continuous performance improvement~\citep{team2025kimi,liu2024let,wang2025dump}.

Within the PCuRL framework, the training process is explicitly structured into three consecutive stages, \ie, easy, medium, and hard. The dataset remains consistent across all three stages. However, in each stage, the difficulty soft weighting mechanism for each stage is employed to preferentially focus on data corresponding to the targeted difficulty level, and a dedicated  $F(\text{Acc})$  function is used to tailor the training focus (as detailed in Section~\ref{sssec:soft_weight_difficulty}). Meanwhile, the data order is independently shuffled to promote sufficient generalization and exploration. This structure ensures the model is exposed to varied learning conditions throughout the curriculum. Notably, the dynamic length reward mechanism is introduced exclusively during the hard stage to further strengthen the model's capacity for complex reasoning. The rationale behind this design is twofold: 1) in the easy and medium stages, the absence of the dynamic length reward allows the model to explore the task space more freely and achieve rapid adaptation and initial performance gains; 2) the challenging questions in the hard stage are the main driving force for model to increases reasoning length and explore the complex reasoning chains. Meanwhile, due to this increased reasoning length brought by the dynamic length reward, the final phase takes more steps for the model to converge than the easy and medium stages.

\section{Experiments}
\label{sec:experiments}

\subsection{Experimental Settings} 
\paragraph{Benchmark datasets.}
To comprehensively assess the performance of \ours, we select a diverse set of challenging multimodal benchmarks spanning multiple domains:
\begin{itemize}
    \item \textbf{Mathematics and Logic Reasoning}: For mathematical problems, we choose the test set of Geometry@3k (Geo3k; \citet{geometry3k}), MathVision~\citep{wang2024measuring} test set, the testmini set of MathVista~\citep{lu2024mathvista}, and MathVerse~\citep{zhang2024mathverse}. For logical reasoning, we use LogicVista~\citep{xiao2024logicvista}. For chart understanding, we utilize the test set of ChartQA~\citep{masry2022chartqa}.
    \item \textbf{Science-related Reasoning}: The test sets of ScienceQA (SciQA; \citet{scienceqa}, MMMU~\citep{yue2024mmmu}, and EMMA~\citep{hao2025can} are selected in our benchmark suite to evaluate the model's capability in the scientific domain.
    \item \textbf{General Understanding}: we further use a general image understanding benchmark, MMStar~\citep{chen2024we}, to measure the model's fundamental visual understanding capability.
\end{itemize}

\paragraph{Baselines.}
We compare \ours against a diverse set of baseline models, encompassing both general-purpose and reasoning-oriented MLLMs of similar model size:
\begin{itemize}
    \item \textbf{General-purpose MLLMs}: Qwen2.5-VL-Instruct-7B~\citep{bai2025qwen25vl}, InternVL2.5-8B~\citep{chen2025internvl25}, InternVL3-8B~\citep{zhu2025internvl3}, and LLaVA-OneVision-7B (LLaVA-OV; \citet{li2024llava}), where are the recent state-of-the-art MLLMs.
    \item \textbf{Reasoning-oriented MLLMs}: MM-Eureka-8B~\citep{meng2025mm}, R1-VL-7B~\citep{zhang2025r1}, MMR1-7B~\citep{leng2025mmr1}, R1-OneVision-7B~\citep{yang2025r1}, OpenVLThinker-7B~\citep{deng2025openvlthinker}, Vision-R1-7B~\citep{huang2025vision}, VL-Rethinker~\citep{wang2025vl}, and ThinkLite-VL-7B~\citep{wang2025sota}.
\end{itemize}

\paragraph{Evaluation.}
We adopt a unified prompt instruction across all evaluations and require models to enclose their final answers within ``\texttt{\textbackslash\text{box\{\}}}'', where the complete prompt is presented in Appendix~\ref{apdx:sys:prompt}. Model inference is conducted using vLLM~\citep{kwon2023efficient} for accelerating generation. For benchmarks that provide official evaluation protocols such as MathVision and MMMU, we strictly adhere to and utilize their respective original evaluation procedures. For other benchmarks, mathematics-related questions are evaluated using Math-Verify\footnote{\url{https://github.com/huggingface/Math-Verify}} and MathRuler\footnote{\url{https://github.com/hiyouga/MathRuler}}, whereas non-mathematical questions are assessed through strict exact matching. To further enhance the robustness and fairness of the evaluation process, two supplementary measures are implemented. First, for multiple-choice questions where the model's generated answer does not correspond to any of the provided options, the most semantically similar option to the generated answer is selected as the final answer. Second, in cases involving open-ended questions where answers cannot be exactly matched or parsers fail to accurately extract the answers, GPT-4o~\citep{openai2024gpt4ocard} is employed as an auxiliary judgment tool.

\paragraph{Implementation Details.}
We utilize the Qwen2.5-VL-Instruct-7B as the backbone model. We train our model using the AdamW optimizer with a learning rate of \texttt{1e-6} under the DeepSpeed-ZeRO3~\citep{rajbhandari2020zero} configuration. For the GRPO configuration, we set the rollout batch size to \texttt{512}, the global batch size to \texttt{128}, and the maximum sequence length to \texttt{4,096}; the KL divergence loss coefficient is configured at \texttt{1e-3}; and \texttt{16} responses are sampled for each prompt with a temperature of \texttt{1.0}. The system prompt used for the training is presented in Appendix~\ref{apdx:sys:prompt}. For the progressive curriculum RL setting, \textbf{our empirical experiments across easy, medium, and hard stages indicate that the reward and validation accuracy are plateaued after approximately 100 optimization steps in the easy and medium stages due to their relatively simple data. Conversely, the hard stage requires substantially more training steps to achieve improved performance}. Thus, we conduct policy optimization for \texttt{100} steps each during both the easy and medium stages, selecting the optimal checkpoint based on the validation performance to serve as the starting point for the subsequent stage. In the hard stage, the model is trained for \texttt{1} epoch (approximately \texttt{200} steps), as incorporating the length reward necessitates more training steps to achieve convergence due to the increased reasoning complexity. For hyperparameters in reward, we set $\alpha = 1$, $\beta = 0.5$, $\gamma = 1$ and $w = 0.25$. The dynamic length reward is considered as punishment, so $r_{\text{len}}^{\text{min}}$ is set to -1 and $r_{\text{len}}^{\text{max}}$ it set to 0. The resultant model is denoted as \ours. For the baseline models, we directly download their open-source versions from HuggingFace\footnote{\url{https://huggingface.co/}} and deploy them under the same environment as ours.

\begin{table}[t]
\centering
\caption{Performance comparison of \ours with other MLLMs on an extensive set of multimodal reasoning benchmarks that encompass mathematical, scientific, and general-domain tasks. All baseline models are reevaluated under identical experimental conditions to ensure a fair comparison; values shown in parentheses denote the results reported in the corresponding original papers. We exclude results for models where benchmark contamination occurred during training, marked with ``-''. The \textbf{bold} and \underline{underline} indicate the best and the second-best scores, respectively.}
\adjustbox{max width=1.0\textwidth}{
\begin{tabular}{l c llllll llll l}
\toprule
\multirow{2}{*}{\textbf{Model}} & \multirow{2}{*}{\textbf{Size}} & \multicolumn{6}{c}{\textbf{Mathematics}} & \multicolumn{4}{c}{\textbf{Science}} & \multicolumn{1}{c}{\textbf{General}}\\
\cmidrule(lr){3-8}\cmidrule(lr){9-12}\cmidrule(lr){13-13}
& & \multicolumn{1}{c}{\textbf{Geo3K}} & \multicolumn{1}{c}{\textbf{MathVerse}} & \multicolumn{1}{c}{\textbf{MathVista}} & \multicolumn{1}{c}{\textbf{MathVision}} & \multicolumn{1}{c}{\textbf{LogicVista}} & \multicolumn{1}{c}{\textbf{ChartQA}} & \multicolumn{1}{c}{\textbf{SciQA}} & \multicolumn{1}{c}{\textbf{MMMU}} & \multicolumn{1}{c}{\textbf{MMMU-Sci}} & \multicolumn{1}{c}{\textbf{EMMA}} & \multicolumn{1}{c}{\textbf{MMStar}} \\
\midrule
\multicolumn{13}{c}{\texttt{General-Purpose Models}}\\
\midrule
\rowcolor[HTML]{F5F5F5} Qwen2.5-VL & 7B & 61.6 & 50.4 (49.2) & 69.3 (68.2) & 28.7 (25.1) & 44.0 & 82.4 & 85.4 & 50.9 & 44.6 & 24.6 & 62.5 (63.9) \\
\rowcolor[HTML]{F5F5F5} InternVL2.5 & 8B & 60.6 & 40.0 (39.5) & 61.4 (64.4) & 19.9 (19.7) & 37.7 (36.0) & 73.4 & 90.3 & 43.1 (48.9)  & 35.0 & 19.9 & 62.2 (62.8)\\
\rowcolor[HTML]{F5F5F5} InternVL3 & 8B & 63.3 & 49.4 (39.8) & 68.5 (71.6) & 30.0 (29.3) & 41.3 (44.1) & 81.3 & 89.3 & 50.8  & 40.6 & 14.5 & 67.5 (68.2) \\
\rowcolor[HTML]{F5F5F5} LLaVA-OV & 7B & 48.5 & 33.6 (26.2) & 56.4 (63.2) & 15.9 & 30.6 & 65.0 & 80.5 & 41.6 & 34.8 & 18.3 & 53.5 (61.7) \\
\midrule
\multicolumn{13}{c}{\texttt{Reasoning-Oriented Models}}\\
\midrule
\rowcolor[HTML]{F5F5F5} MM-Eureka & 8B & 67.2 & 52.3 (50.3) & 73.4 (73.0) & 29.4 (26.9) & \underline{47.1} & 82.7 & 86.4 & 52.3 & 46.7 & 27.4 & 64.7 \\
\rowcolor[HTML]{F5F5F5} R1-VL & 7B & 57.5 & 41.3 (40.0) & 61.5 (63.5) & 23.0 (24.7) & 36.3 & 76.3 & 86.0 & 38.1 & 33.4 & 24.0 & 55.6 (60.0)\\
\rowcolor[HTML]{F5F5F5} MMR1 & 7B & 65.9 & 52.5 (45.1) & 73.6 (71.0) & \textbf{32.9} (30.2) & 46.6 (50.8) & 82.8 & \underline{86.7} & \textbf{53.1} & \textbf{47.3} & 28.1 & \textbf{66.3}\\
\rowcolor[HTML]{F5F5F5} R1-OneVision & 7B & 57.9 & 44.0 (46.4) & 60.3 (64.1) & 22.0 (29.9) & 40.0 & 72.5 & 85.3 & 43.4 & 37.3 & 22.2 & 56.2 \\
\rowcolor[HTML]{F5F5F5} OpenVLThinker & 7B & 60.6 & 48.1 (47.9) & 70.6 (70.2) & 22.0 (25.3) & 41.0 & 81.0 & 85.9 & 50.9 & 43.9 & 24.9 & 62.8 \\
\rowcolor[HTML]{F5F5F5} VL-Rethinker & 7B & \underline{67.7} & \textbf{54.6} (54.2) & \underline{73.7} (74.9) & 30.1 (32.3) & 45.7 & \textbf{83.5} & \underline{86.7} & \underline{52.9} & 46.5 & \underline{28.6} (29.7) & 64.2 \\
\rowcolor[HTML]{F5F5F5} Vision-R1 & 7B & 67.0 & 51.9 (52.4) & 72.1 (73.5) & - & 44.7 & 82.7 & - & 26.4 & 26.3 & 28.3 & \underline{65.4} \\
\rowcolor[HTML]{F5F5F5} ThinkLite-VL & 7B & 63.5 & 51.3 (50.7) & 72.5 (75.1) & 27.5 & 44.3 & 83.1 & - & 50.9 & 44.3 & 26.4 & 64.6 (65.0) \\
\midrule
\rowcolor[HTML]{E8F5E9} \ours & 7B & \textbf{68.7} & \underline{53.3} & \textbf{74.8} & \underline{30.7} & \textbf{48.9} & \underline{83.4} & \textbf{87.6} & 52.6 & \underline{47.0} & \textbf{29.1} & \textbf{66.3} \\
\bottomrule
\end{tabular}}
\label{tab:mm_result}
\end{table}

\subsection{Performance Comparison on Diverse Multimodal Benchmarks}
Table~\ref{tab:mm_result} presents a detailed comparison between \ours and a diverse set of both general-purpose and reasoning-oriented MLLMs across ten different multimodal benchmarks. General-purpose models, such as the InternVL series and Qwen2.5-VL, demonstrate strong performance in general and certain scientific domains. However, they typically underperform compared to reasoning-oriented MLLMs on mathematical benchmarks, which demand advanced analytical and reasoning capabilities to reach correct answers. In comparison to the backbone model Qwen2.5-VL-Instruct, \ours exhibits substantial performance enhancements across all evaluated benchmarks, encompassing mathematics, science, and general domains. \ours achieves absolute gains of $7.6\%$, $5.5\%$, and $4.9\%$ on Geometry@3K, MathVista, and LogicVista, respectively, as well as improvements of $2.2\%$ on ScienceQA, $4.5\%$ on EMMA, and $3.8\%$ on MMStar. These results clearly demonstrate that our approach yields consistent and robust improvements across a diverse range of domains, rather than being tailored solely to a particular domain.

Among reasoning-oriented models, \ours attains either the best or highly competitive performance \textit{even without necessitating a cold-start warm-up phase}. This result underscores the effectiveness of our proposed progressive curriculum learning strategy. Notably, it achieves the highest performance on 6 out of 10 multimodal benchmarks, demonstrating exceptional capability on rigorous mathematics and scientific benchmarks, while also attaining superior results on comparatively less complex tasks such as ScienceQA and MMStar. Compared to VL-Rethinker, \ours achieves marginally lower performance only on MathVerse, ChartQA, and MMMU, while outperforming it across other benchmarks. VL-Rethinker adopts a forced rethinking strategy, which explicitly directs the model to engage in multiple iterations of reasoning during RL training. In contrast, \ours exclusively leverages the model's capacity for autonomous exploration, without relying on additional guidance. Although models such as R1-VL, R1-OneVision, OpenVLThinker, and Vision-R1 utilize cold-start initialization with meticulously curated SFT datasets, \ours is consistently superior to these approaches. These findings underscore the efficacy of our RL pipeline in enhancing the model's reasoning capabilities across a diverse range of domains.

\subsection{Ablation Study}
In this section, we conduct a comprehensive ablation study to analyze the effectiveness of each component within PCuRL, including the multi-stage curriculum strategy, online difficulty soft weighting, and dynamic length reward. Furthermore, we examine the training dynamics by visualizing key aspects of the reinforcement learning process, including reward trajectories and length propagation curves, to provide deeper insights into the model's learning behaviors.

\begin{table}[t]
\centering
\caption{The component-wise performance decomposition of the PCuRL framework, where ``+Curriculum'' and ``+DyLR'' represent the addition of progressive curriculum strategy and dynamic length reward to GRPO, respectively. The \textbf{bold} and \underline{underline} indicate the best and the second-best scores, respectively.}
\adjustbox{max width=1.0\textwidth}{
\begin{tabular}{l cccccc ccc c c}
\toprule
\multirow{2}{*}{\textbf{Model}} & \multicolumn{6}{c}{\textbf{Mathematics}} & \multicolumn{3}{c}{\textbf{Science}} & \multicolumn{1}{c}{\textbf{General}} & \multirow{2}{*}{\textbf{Avg}}\\
\cmidrule(lr){2-7}\cmidrule(lr){8-10}\cmidrule(lr){11-11}
& \textbf{Geo3K} & \textbf{MathVerse} & \textbf{MathVista} & \textbf{MathVision} & \textbf{LogicVista} & \textbf{ChartQA} & \textbf{SciQA} & \textbf{MMMU} & \textbf{EMMA} & \textbf{MMStar} & \\
\midrule
\rowcolor[HTML]{F5F5F5} Vanilla GRPO & 66.1 & 52.2 & 71.4 & 30.0 & 44.0 & \underline{83.8} & \textbf{87.8} & 51.4 & 28.0 & \textbf{66.3} & 58.1\\
\rowcolor[HTML]{F5F5F5} +Curriculum & \underline{67.4} & 51.9 & \underline{74.0} & \underline{30.4} & 47.5 & \textbf{83.9} & \underline{87.6} & \textbf{52.7} & \underline{28.3} & \underline{65.2} & \underline{58.9}\\
\rowcolor[HTML]{F5F5F5} +DyLR & 66.2 & \underline{52.5} & 73.3 & 29.4 & \underline{48.4} & 83.2 & 87.1 & \underline{52.6} & 28.2 & 64.9 & 58.6\\
\rowcolor[HTML]{E8F5E9} \ours & \textbf{68.7} & \textbf{53.3} & \textbf{74.8} & \textbf{30.7} & \textbf{48.9} & 83.4 & \underline{87.6} & \underline{52.6} & \textbf{29.1} & \textbf{66.3} & \textbf{59.5}\\
\bottomrule
\end{tabular}}
\label{tab:ablation}
\end{table}

\begin{table}[t]
\centering
\caption{Ablation study examining the Online Difficulty Soft weighting (ODSW) in our progressive curriculum RL settings. ``Binary'' denotes the binary weighting strategy, where we assess three difficulty ranges of $[T_{\text{min}}, T_{\text{max}}]$; ODSW ``Easy'', ``Medium'', and ``Hard'' represent only utilizing the three ODSW variants during the RL training, respectively. The \textbf{bold} and \underline{underline} indicate the best and the second-best scores.}
\adjustbox{max width=1.0\textwidth}{
\begin{tabular}{l cccccc ccc c c}
\toprule
\multirow{2}{*}{\textbf{Model}} & \multicolumn{6}{c}{\textbf{Mathematics}} & \multicolumn{3}{c}{\textbf{Science}} & \multicolumn{1}{c}{\textbf{General}} & \multirow{2}{*}{\textbf{Avg}}\\
\cmidrule(lr){2-7}\cmidrule(lr){8-10}\cmidrule(lr){11-11}
& \textbf{Geo3K} & \textbf{MathVerse} & \textbf{MathVista} & \textbf{MathVision} & \textbf{LogicVista} & \textbf{ChartQA} & \textbf{ScienceQA} & \textbf{MMMU}  & \textbf{EMMA} & \textbf{MMStar} \\
\midrule
\rowcolor[HTML]{F5F5F5} Binary $[0.50,1.00]$ & 59.6 & 49.3 & 72.0 & 27.8 & 41.5 & 82.6 & 87.1 & 49.1 & 25.2 & 64.5 & 55.9 \\
\rowcolor[HTML]{F5F5F5} Binary $[0.25,0.75]$ & 67.2 & 51.9 & 72.3 & \underline{30.4} & 46.7 & \textbf{84.7} & 87.6 & 51.0 & 27.0 & 64.4 & 58.3 \\
\rowcolor[HTML]{F5F5F5} Binary $[0.00,0.50]$ & 65.6 & 51.4 & 73.1 & 29.5 & 45.3 & 83.9 & 86.4 & 51.8 & 25.5 & \underline{65.4} & 57.8 \\
\midrule
\rowcolor[HTML]{F5F5F5} ODSW Easy only & 64.2 & 51.8 & 72.4 & 28.6 & \underline{48.2} & 83.4 & 87.3 & 51.3 & \underline{28.2} & 64.1 & 58.0 \\
\rowcolor[HTML]{F5F5F5} ODSW Medium only & \underline{68.2} & 51.9 & \underline{74.7} & 29.9 & 44.2 & 83.9 & \textbf{88.3} & 52.4 & 27.9 &  64.8 & 58.6 \\
\rowcolor[HTML]{F5F5F5} ODSW Hard only & \underline{68.2} & \underline{52.3} & 74.5 & 29.5 & 46.2 & \underline{84.0} & \underline{88.0} & \textbf{52.7} & 27.0 & 64.6 & \underline{58.7} \\
\midrule
\rowcolor[HTML]{E8F5E9} \ours & \textbf{68.7} & \textbf{53.3} & \textbf{74.8} & \textbf{30.7} & \textbf{48.9} & 83.4 & 87.6 & \underline{52.6} & \textbf{29.1} & \textbf{66.3} & \textbf{59.5} \\
\bottomrule
\end{tabular}}
\label{tab:pattern}
\end{table}

\subsubsection{Component-wise Performance Decomposition of PCuRL}
Table~\ref{tab:ablation} presents a systematic ablation that quantifies the marginal contribution of each PCuRL component. Compared to the vanilla GRPO baseline, \ours enhances the performance on the more demanding benchmarks,\ie, MathVision, LogicVista, MMMU, and EMMA, while preserving comparable results on the easier suites, \ie, ChartQA, ScienceQA, and MMStar, confirming that our pipeline is more beneficial where reasoning depth is critical.

Introducing the multi-stage progressive curriculum alone, \ie, ``+Curriculum,'' yields consistent gains across nearly every task, underscoring the value of an easy-to-hard progression. Appending the dynamic length reward at the final curriculum stage further elevates scores on the most challenging mathematical datasets, \eg, MathVista and LogicVista, pushing the overall average score to $59.5\%$. In contrast, when the dynamic length reward is applied directly, \ie, ``+DyLR,'' to the GRPO baseline—without the stabilizing effect of progressive curriculum learning—the model exhibits unstable and inconsistent performance. We suppose that, during the early phases of RL training, many format-mismatched or ill-structured queries are incorrectly interpreted by the model as inherently ``difficult'' due to their lack of correct responses. As a result, DyLR prematurely encourages the model to extend its reasoning length excessively in response to these misclassified inputs, which not only leads to inefficient learning but also disrupts the optimization trajectory, ultimately hindering convergence and generalization.

\subsubsection{Impact of Online Difficulty Soft Weighting on Model Performance}
To assess how various difficulty weighting methods influence the effectiveness of \ours, we conduct a systematic ablation using single-stage RL training with different weighting strategies, and the results are summarized in Table~\ref{tab:pattern}.

\textbf{Soft weighting consistently surpasses binary weighting.}
The results indicate that online difficulty soft weighting (ODSW) achieves superior performance relative to binary weighting schemes across the majority of benchmarks. Notably, when target difficulty is highly skewed or imbalanced, the advantages of the soft weighting approach become particularly pronounced. This suggests that soft weighting maintains a broader and more balanced optimization direction by dynamically emphasizing appropriate difficulty levels. Conversely, excessively focusing on either solely easy or hard samples tends to degrade the overall reasoning capability. Consequently, we adopt online difficulty soft weighting within our multi-stage progressive curriculum RL framework to sustain comprehensive model development.

\textbf{Exposure to challenging queries is pivotal for deep reasoning gains.}
Our analysis further reveals that models receiving disproportionately more training on easier samples generally exhibit suboptimal performance. In contrast, placing greater emphasis on more challenging problems—especially through a soft weighting approach—consistently improves outcomes, underscoring the essential role of difficult questions in advancing reasoning ability. This insight motivates our design choice to progressively prioritize challenging tasks during the final stages of curriculum RL training, thereby maximizing the model's reasoning proficiency.

\begin{figure}[t]
    \centering
    \includegraphics[width=0.8\textwidth]{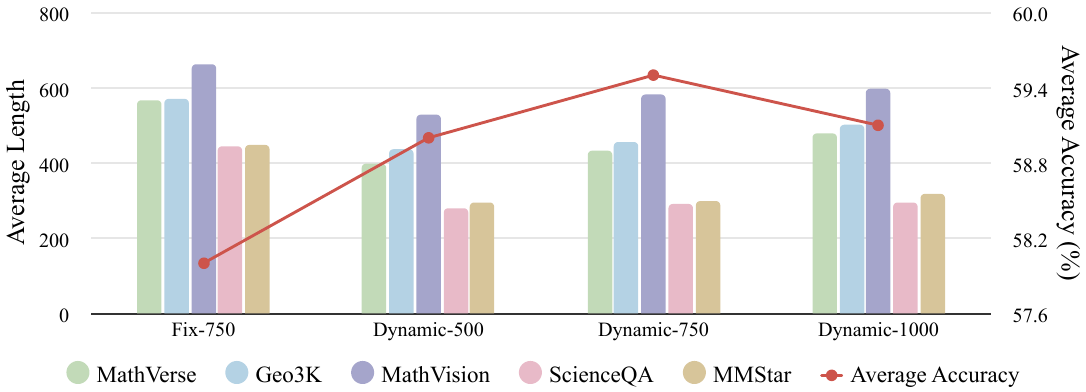}
    \caption{Performance comparison of models trained with different length reward strategies. ``Dynamic-$N$'' denotes models employing our dynamic length reward with a target length of $N$ during the final stage of curriculum RL. ``Fix-$N$'' refers to models trained with a fixed-length reward that enforces the fixed target length of $N$ across all responses. We visualize both the average response length and the overall accuracy across selected benchmarks.}
    \label{fig:len}
\end{figure}

\subsubsection{Impact of Dynamic Length Reward on Reasoning Efficiency and Performance}
Figure~\ref{fig:len} illustrates how different length-reward strategies influence the average reasoning accuracy of \ours, alongside the corresponding average token lengths of generated responses on representative multimodal benchmarks. Specifically, we compare the performance of our dynamic length reward approach, with varying target lengths, against the fixed-length cosine reward method proposed by \citet{yeo2025demystifying}. The fixed-length method uniformly encourages all incorrect outputs to reach a pre-determined target length, regardless of the actual complexity or demands of the questions. We denote our dynamic length reward as Dynamic-$N$, where $N\in\{500, 750, 1000\}$, and the fixed-length cosine reward as Fix-$N$, where we set $N=750$.

\textbf{Dynamic length reward consistently surpasses fixed-length reward.} 
Our findings reveal that models guided by dynamic-length rewards achieve noticeably higher average accuracy across benchmarks, coupled with a more adaptive and context-appropriate response length distribution. Dynamic reward mechanisms tend to generate shorter, more concise reasoning paths overall, while exhibiting significant flexibility—longer responses for difficult problems and shorter responses for simpler ones. In contrast, fixed-length rewards indiscriminately inflate response lengths, often imposing unnecessary complexity and redundant reasoning on relatively straightforward tasks. The inferior accuracy scores further confirm that mere increases in reasoning length do not necessarily translate into enhanced performance in multimodal reasoning tasks.

\textbf{Dynamic reward selectively lengthens responses primarily for challenging tasks.}
Further investigation into the influence of target lengths highlights a clear trend: as we elevate the target reasoning length (from $500$ to $1,000$ tokens), response lengths on mathematics-related datasets grow steadily, reflecting the inherent complexity and depth requirements of these tasks. While benchmarks requiring less reasoning, such as ScienceQA and MMStar, remain unaffected and maintain concise responses. This pattern confirms our expectation that dynamic length rewards strategically and beneficially extend response lengths specifically for challenging queries, thereby optimizing reasoning depth precisely where it matters most.

\begin{figure}[t]
    \centering
    \begin{subfigure}[t]{0.32\textwidth}
        \centering
        \includegraphics[width=\textwidth]{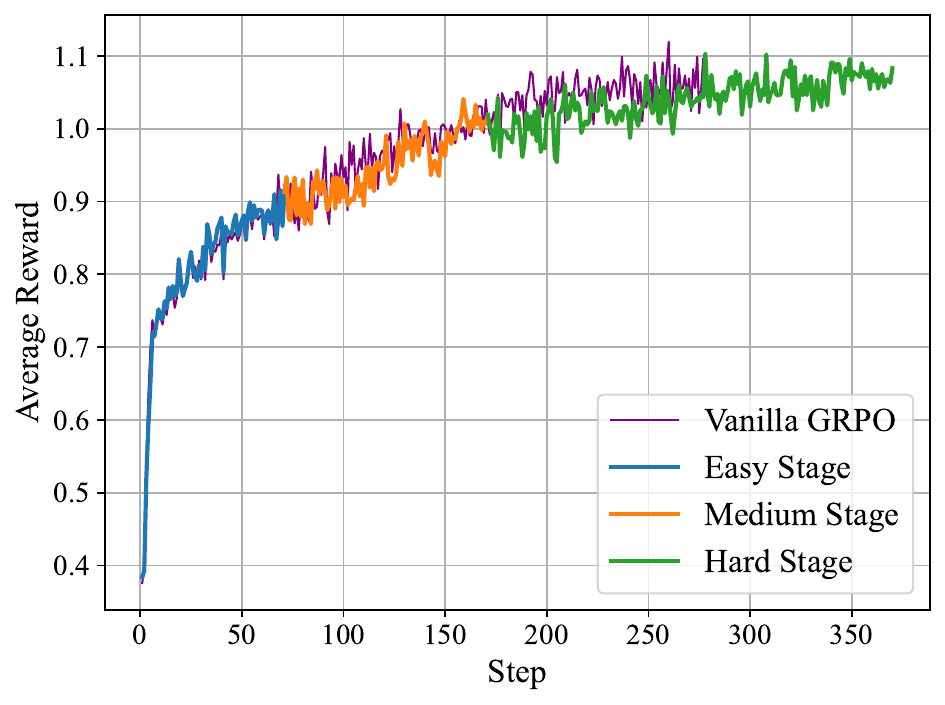}
        \caption{Average Reward Curve}
        \label{fig:curve_reward}
    \end{subfigure}
    \hfill
    \begin{subfigure}[t]{0.32\textwidth}
        \centering
        \includegraphics[width=\textwidth]{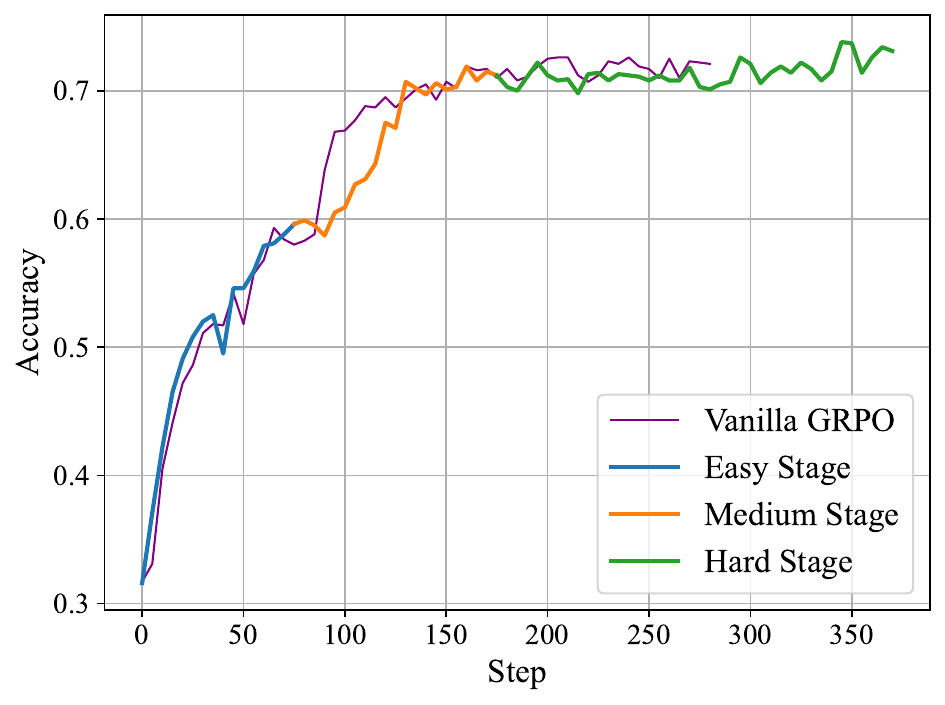}
        \caption{Validation Accuracy Curve}
        \label{fig:curve_val}
    \end{subfigure}
    \hfill
    \begin{subfigure}[t]{0.32\textwidth}
        \centering
        \includegraphics[width=\textwidth]{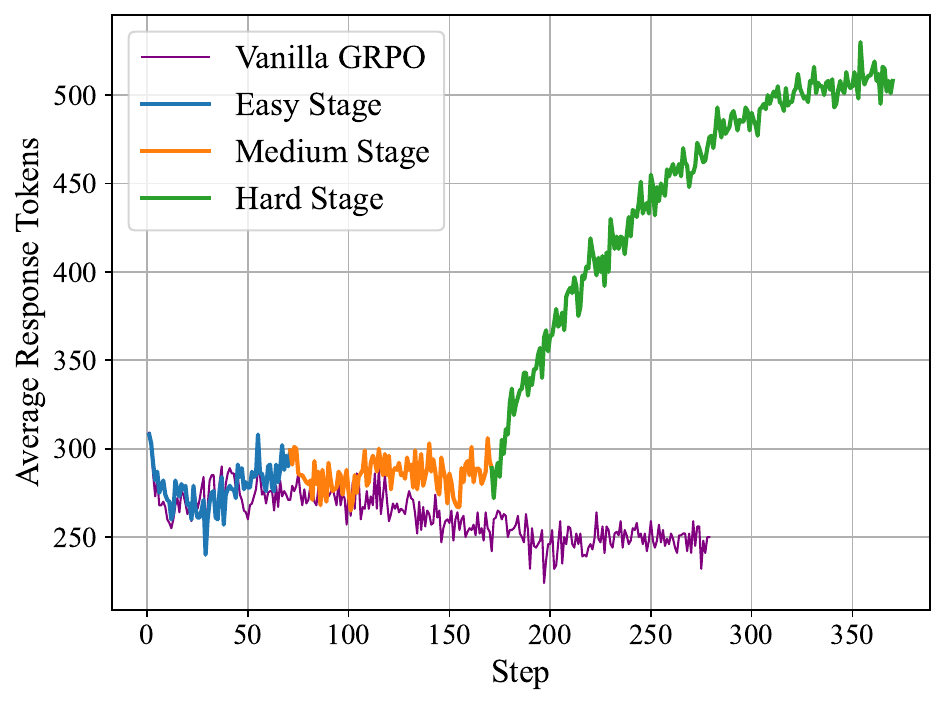}
        \caption{Average Response Length Curve}
        \label{fig:curve_len}
    \end{subfigure}
    \caption{Training curves for PCuRL (with a target response length of $500$ tokens) and vanilla GRPO. The average reward curve indicates the mean reward of sampled responses during training. The validation accuracy curve shows model performance on a held-out validation set (around $1,000$ questions, split from the original training set at initialization) as measured by the accuracy reward function. The average length curve displays the mean response length of sampled outputs during training.}
    \label{fig:curve}
\end{figure}

\subsubsection{Comparison of Training Dynamics between PCuRL and Vanilla GRPO}
Figure~\ref{fig:curve_reward} demonstrates that, during its initial stages, PCuRL's reward trajectory closely aligns with that of vanilla GRPO, suggesting similar reward functions focused primarily on correctness. Both the easy and medium stages exhibit steadily increasing rewards. At the onset of the hard stage, PCuRL's average reward drops slightly below the baseline due to the dynamic length reward, which penalizes responses beneath the growing target length. The model initially struggles to meet this expanded requirement, temporarily reducing its average reward. However, as it learns to generate longer responses, the reward recovers, eventually matching or exceeding that of the vanilla model.

\begin{figure}[t]
    \centering
    \includegraphics[trim={8mm 1cm 8mm 1.2cm},clip,width=0.85\textwidth]{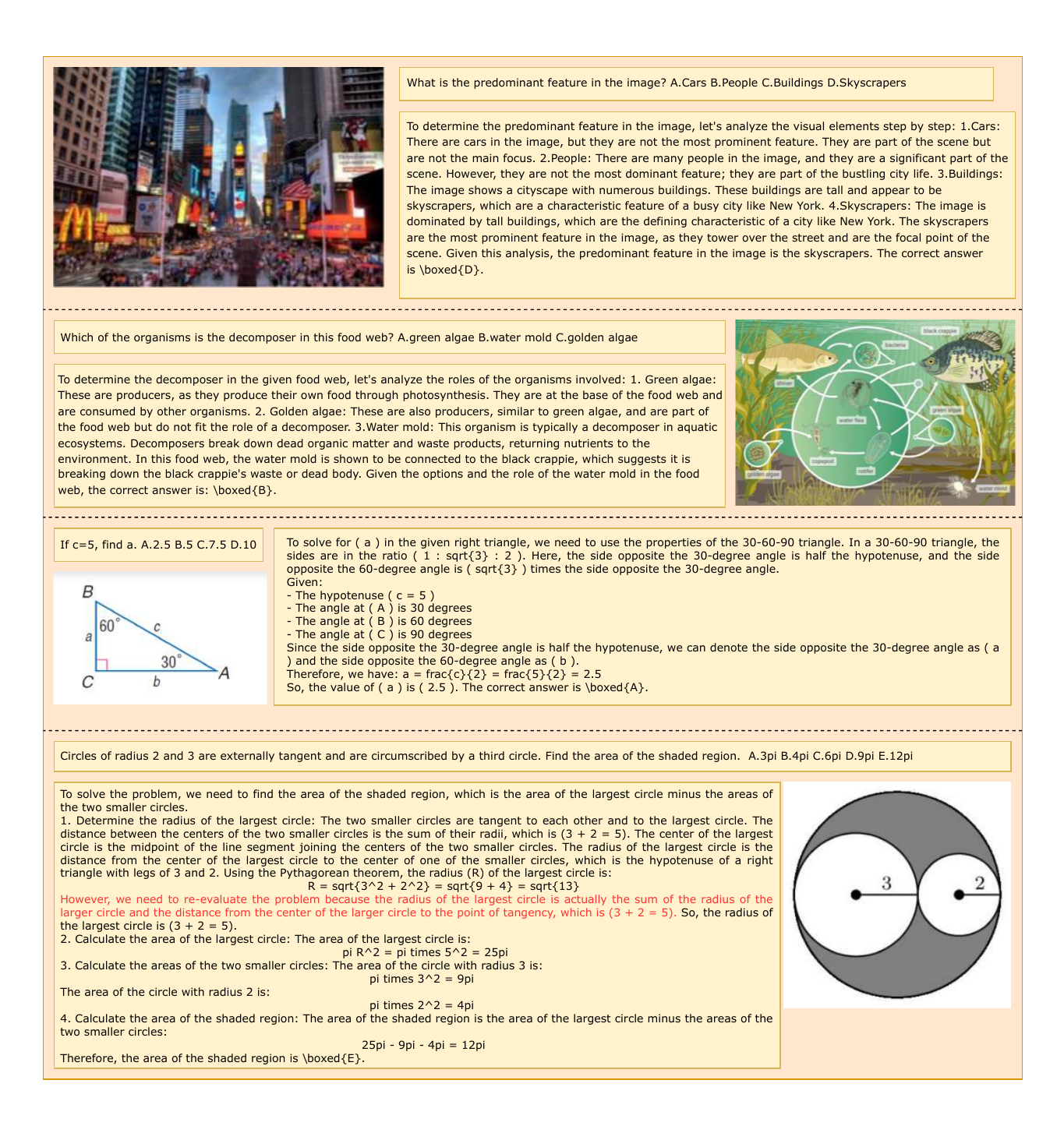}
    \caption{Case studies of \ours, where samples are drawn from multiple benchmarks, including MMStar, ScienceQA, Geometry@3K, and MathVision.}
    \label{fig:case}
\end{figure}

Figure~\ref{fig:curve_val} illustrates the corresponding validation accuracy. In the easy and medium stages, PCuRL's accuracy mirrors that of vanilla GRPO, both rising from $0.3$ to around $0.7$. As depicted in Figure~\ref{fig:curve_len}, the hard stage introduces a significant increase in response length, coinciding with PCuRL's validation accuracy surpassing the baseline. While vanilla GRPO's accuracy saturates at approximately $0.7$, PCuRL continues to improve, reaching a new peak. This result suggests that the extended reasoning chains produced during the hard stage effectively enhance problem-solving performance, rather than merely increasing output length.

Figure~\ref{fig:curve_len} displays the average response length for both methods. Vanilla GRPO's response length remains relatively static, fluctuating between $250$ and $300$ tokens, and even exhibits a slight downward trend, highlighting its limitation in encouraging diverse or extended reasoning. In contrast, PCuRL effectively modulates response length according to its curriculum: during the easy and medium stages, lengths remain similar to the baseline, indicating a focus on correctness. In the hard stage, the introduction of the dynamic length reward drives a marked increase in response length, rising consistently from approximately $280$ to the target of $500$ tokens. This demonstrates the curriculum's capacity to guide the model toward more sophisticated outputs incrementally.

In summary, these findings indicate that the progressive curriculum adopted in PCuRL is highly effective in training MLLMs to generate longer and more intricate reasoning. Unlike standard GRPO, which neglects response length, PCuRL's use of a dynamic length reward in the hard stage not only achieves the desired target length but also leads to notable gains in validation accuracy. The observed reward and accuracy curves collectively confirm the successful adaptation of the model to evolving objectives throughout the curriculum.

\subsection{Case Study}
Figure~\ref{fig:case} illustrates multimodal reasoning cases of \ours across diverse domains:

\textbf{Reasoning Trajectories}: For multiple-choice questions in both general and scientific domains, \ours systematically evaluates each option to identify the correct answer. In contrast, when addressing mathematical problems, the model typically first derives the solution independently and subsequently maps it to the given options. The third and fourth examples further exemplify the model's capacity to handle math questions of varying complexity. In the comparatively straightforward third case, the model promptly recalls properties of a 30-60-90 triangle and directly provides the solution. Conversely, in the more challenging fourth case, the model adopts a meticulous, stepwise approach, decomposing the problem into four discrete reasoning steps. These examples collectively demonstrate that the model dynamically adjusts its reasoning depth, employing more extensive and detailed reasoning processes when confronted with more complex problems.

\textbf{Self-Reflective Reasoning}: Notably, the model also exhibits self-reflection capabilities, as evidenced in the fourth example (highlighted in \textcolor{red}{red} in the figure). Initially, the model incorrectly applied the Pythagorean theorem when calculating the radius of the largest circle, leading to an erroneous result. However, it promptly recognized and addressed this error by invoking the ``re-evaluate'' mechanism, subsequently correcting its calculation and ensuring the accuracy of the remaining steps. This behavior underscores the effectiveness of our RL-based training pipeline in instilling valuable self-reflective abilities within multimodal models.

\section{Conclusion}
\label{sec:conclusion}
In this work, we introduce \ours, an advanced multimodal large language model (MLLM) enhanced by a progressive curriculum-based reinforcement learning framework termed PCuRL, designed to systematically improve multimodal reasoning capabilities. The multi-stage curriculum embedded within PCuRL progressively guides the model on tasks of increasing complexity, enhancing its proficiency in addressing reasoning challenges across various domains. A key component of our framework is an online difficulty soft weighting mechanism, which dynamically adjusts task difficulty in response to the model's evolving abilities, facilitating a balanced transition from simpler to more complex tasks. Furthermore, the dynamic length reward mechanism modulates the length of the model's responses according to problem-specific demands, optimizing the balance between reasoning depth and efficiency. Experimental results demonstrate that \ours achieves the state-of-the-art or highly competitive performance across various benchmarks, underscoring the substantial potential of meticulously crafted curriculum learning strategies to broaden the applicability of multimodal reasoning models.

\bibliographystyle{assets/plainnat}
\bibliography{paper}

\clearpage
\newpage
\beginappendix

\section{Data Statistics}
\label{apdx:sec:data}
Table~\ref{tab:datasets} summarizes the distribution of RL training data, where we show both the number of samples and the filter rate per dataset.

\begin{table}[t]
\centering
\caption{Data statistics of the training dataset in the RL stage, where the filter rate refers to the percentage of questions removed in the difficulty sampling, and the data size represents the number of samples after sampling.}
\footnotesize
\begin{tabular}{lcccc}
\toprule
\textbf{Category} & \textbf{QA Type} & \textbf{Dataset} & \textbf{Data size} & \textbf{Filter rate} \\ 
\midrule
\multirow{6}{*}{Mathematics} & Open-ended & Geometry3K & 2101 & 39\% \\ 
 \cmidrule(lr){2-5} 
 & Open-ended & GeoQA+ & 7886 & 50\% \\ 
 \cmidrule(lr){2-5} 
 & Open-ended & Geos & 367 & 22\% \\ 
 \cmidrule(lr){2-5} 
 & Open-ended & GeomVerse & 8179 & 12\% \\ 
 \cmidrule(lr){2-5} 
 & Open-ended & Inter-GPS & 946 & 26\% \\ 
 \cmidrule(lr){2-5} 
 & Open-ended & MultiMath & 13503 & 35\% \\ 
 \midrule
\multirow{3}{*}{Logical} & Open-ended & Raven & 6919 & 65\% \\ 
\cmidrule(lr){2-5} 
 & Open-ended & MM-IQ & 2492 & 7\% \\ 
 \cmidrule(lr){2-5} 
 & Open-ended & EasyArc & 600 & 0\%\\ 
 \midrule
 \multirow{2}{*}{Counting} & Open-ended & CLEVR-Math & 1000 & 92\% \\ 
 \cmidrule(lr){2-5} 
 & Open-ended & Super-CLEVR & 3000 & 20\% \\ 
 \midrule
 \multirow{3}{*}{Science} & Open-ended & AI2D & 5034 & 35\%\\ 
 \cmidrule(lr){2-5} 
 & Multi-choice & ScienceQA & 1098 & 82\% \\ 
 \cmidrule(lr){2-5} 
 & Open-ended & TQA & 5959 & 31\% \\ 
 \midrule
 \multirow{6}{*}{Charts} & Open-ended & ChartQA & 5570 & 72\% \\ 
 \cmidrule(lr){2-5} 
 & Open-ended & TabMWP & 2963 & 70\% \\ 
 \cmidrule(lr){2-5} 
 & Open-ended & DVQA & 2589 & 45\% \\ 
 \cmidrule(lr){2-5}
 & Open-ended & FigureQA & 2000 & 40\% \\ 
 \cmidrule(lr){2-5}
 & Open-ended & ArXivQA & 2000 & 57\% \\ 
 \cmidrule(lr){2-5}
 & Open-ended & InfographicVQA & 626 & 70\%\\ 
 \midrule
\multirow{3}{*}{General} & Open-ended & OKVQA & 1500 & 12\% \\ 
\cmidrule(lr){2-5} 
 & Open-ended & VQA2.0 & 1500 & 22\% \\ 
 \cmidrule(lr){2-5} 
 & Open-ended & LLaVA-CoT & 1500 & 8\% \\ 
 \bottomrule
\end{tabular}
\label{tab:datasets}
\end{table}

\section{System Prompt}
\label{apdx:sys:prompt}
\textbf{Training Prompt:} We use this system prompt in RL training to encourage the model to follow the reasoning format and put the final answer in a ``\textbackslash boxed\{\}''. 

\begin{table}[ht]
\caption{The system prompt used for RL training.}
\label{tab:rl_prompt}
\begin{tcolorbox}[colframe=orange!60, colback=orange!20, fonttitle=\bfseries\large, coltitle=black, boxrule=0.75mm, arc=5mm, auto outer arc, width=\textwidth,toptitle=6pt, bottomtitle=6pt]
\small
\setstretch{1.2}
A conversation between User and Assistant.\\
The User provides an image and asks a question.
The Assistant first analyzes both the image and the question, then carefully thinks about the reasoning process step by step, and finally provides the User with an accurate answer.
The Assistant must carefully checkout the correctness and validity of each reasoning step.
If any errors or inconsistencies are found during the reasoning process, the Assistant reflects and corrects them logically. \\
The reasoning process and answer are enclosed within <think> </think> and <answer> </answer> tags, respectively, i.e., <think> detailed reasoning process here, with potential reflections and corrections </think><answer> final answer here, with the key result enclosed within \textbackslash boxed\{\} </answer>
\end{tcolorbox}
\end{table}

\textbf{Evaluation Prompt:} To ensure fairness during the evaluation, we adopt a simple prompt that can be generalized to most models rather than the system prompt used in the training. 

\begin{table}[ht]
\caption{The system prompt used for evaluation.}
\label{tab:eval_prompt}
\begin{tcolorbox}[colframe=orange!60, colback=orange!20, fonttitle=\bfseries\large, coltitle=black, boxrule=0.75mm, arc=5mm, auto outer arc, width=\textwidth,toptitle=6pt, bottomtitle=6pt]
\small
\setstretch{1.2} %
Please solve the problem step by step and put your answer in one ``\textbackslash boxed\{\}''. If it is a multiple-choice question, only one letter is allowed in the ``\textbackslash boxed\{\}''.
\end{tcolorbox}
\end{table}

\end{document}